\documentclass{ieeeaccess}
\usepackage{cite}
\usepackage{graphicx}
\usepackage{float}
\usepackage{amsmath,amssymb,amsfonts}
\usepackage{algorithmic}
\usepackage{graphicx}
\usepackage{textcomp}

\usepackage{booktabs}

\usepackage{booktabs}
\def\BibTeX{{\rm B\kern-.05em{\sc i\kern-.025em b}\kern-.08em
    T\kern-.1667em\lower.7ex\hbox{E}\kern-.125emX}}
\begin{document}
\history{}
\doi{}

\title{Semi-supervised Object Detection: A Survey on Recent Research and Progress}
\author{\uppercase{Yanyang Wang},
\uppercase{Zhaoxiang Liu}, \uppercase {and Shiguo Lian}}

\address{Unicom Digital Technology, China Unicom, Beijing 100013, China}

\corresp{Corresponding author: Zhaoxiang Liu (e-mail: liuzx178@chinaunicom.cn), Shiguo Lian (e-mail: liansg@chinaunicom.cn)}

\begin{abstract}
In recent years, deep learning technology has been maturely applied in the field of object detection, and most algorithms tend to be supervised learning. However, a large amount of labeled data requires high costs of human resources, which brings about low efficiency and limitations. Semi-supervised object detection (SSOD) has been paid more and more attentions due to its high research value and practicability. It is designed to learn information by using small amounts of labeled data and large amounts of unlabeled data. In this paper, we present a comprehensive and up-to-date survey on the SSOD approaches from five aspects. We first briefly introduce several ways of data augmentation. Then, we dive the mainstream semi-supervised strategies into pseudo labels, consistent regularization, graph based and transfer learning based methods, and introduce some methods in challenging settings. We further present widely-used loss functions, and then we outline the common benchmark datasets and compare the accuracy among different representative approaches. Finally, we conclude this paper and present some promising research directions for the future. Our survey aims to provide researchers and practitioners new to the field as well as more advanced readers with a solid understanding of the main approaches developed over the past few years.
\end{abstract}


\titlepgskip=-21pt

\maketitle

\section{Introduction}
\label{sec:introduction}
\PARstart{I}{n} recent years, deep learning techniques have achieved great results in data mining, computer vision, natural language processing, multimedia learning, etc. Object detection is one of the most important and challenging branches in computer vision, and it has been widely used in people's lives, such as surveillance security and automatic driving. With the rapid development of deep learning networks for detection tasks, the performance of object detectors has been greatly improved. To perform the detection tasks, deep learning techniques accomplish object localization and classification by extracting and analyzing features of sample data, so labeled samples provide a large data base for object detection tasks and large dataset such as MS-COCO [3] has largely triggered the boom of deep learning in object detection.

However, in many real-world object detection tasks, the labeled samples need a lot of manual labeling and thus require a lot of human and financial resources, so it is difficult to obtain sufficient labeled data in some cases. All of the above reasons lead to samples with few labels and even fewer high quality labels. Besides that, a small amount of data can not satisfy the training requirements of the model, and the accuracy and robustness of the model will be affected. Therefore, during the training process, the information of a large number of unlabeled samples needs to be reasonably used, which can effectively improve the task performance. Semi-supervised learning makes a great contribution in this direction and a large body of work have investigated the application of semi-supervised techniques to object detection tasks.

However, to our best knowledge, no review has been done for the recently emerged semi-supervised learning based object detection methods. This survey will provide a comprehensive review of the state-of-the-art (SOTA) algorithms for semi-supervised object detection, whose innovation points and frameworks are described in detail. A SSOD system generally consists of three modules: data augmentation, semi-supervised strategy and loss function. We also construct taxonomy from the process of SSOD to overview all these approaches(as shown in Fig. 1).The subsequent content of the article is organized as follows: related work in Section II, introduction to data augmentation in Section III, semi-supervised strategies in Section IV,  challenging settings in Section V, overview of loss in Section VI, datasets and comparison results in Section VII, conclusion and future directions in Section VIII.

\section{Related Work}
Semi-supervised learning is a learning method that combines supervised learning with unsupervised learning and focuses on how to make reasonable use of labeled and unlabeled samples in the training process [4] [5]. It establishes the relationship between predicted samples and learning objectives by applying some hypotheses: (1) The smoothness assumption assumes that two samples are more likely to have the same class labels when they are closely located in a high-density region. (2) The cluster assumption assumes that when two samples are in the same cluster, they are very likely to be of the same class. (3) The manifold assumption assumes that two samples have similar class labels when they lie within a small local neighborhood in the low-dimensional manifold. 

Based on these assumptions, there are many well-established algorithms often applied to image classification tasks, and the papers [4]-[8] provide comprehensive surveys to these algorithms. These semi-supervised learning methods for image classification can be classified into the following categories: generative methods [9] [10]; graph-based methods [11] [12]; consistency regularization methods [13] [14]; pseudo-label methods [15]-[18] and hybrid methods [19]-[21].

There have been already some reviews doing research on supervised learning object detection. Specifically, the paper [1] provides a comprehensive survey of many aspects of generic object detection research: leading detection frameworks and fundamental subproblems including object feature representation, object proposal generation, context information modeling, training strategies and evaluation issues. The paper [2] lists the traditional and new applications, and also introduces representative branches of object detection. The paper [3] provides an in-depth analysis of major object detectors in both categories-one and two stage detectors, and also presents a detailed evaluation of the landmark backbone architectures and lightweight models.

Though there are already some surveys about semi-supervised image classification and supervised object detection, as far as we know, our paper is the first review on semi-supervised object detection.

\section{Data augmentation}
Data augmentation is critical to improve model generalization and robustness, which is the first step in SSOD. In order to improve the robustness of the model and make reasonable use of unlabeled data information, consistency regularization is used to constraint the augmented data for the consistency of output labels. The ways of augmentation are quite different because of the discrepancy between different ways.

\subsection{Strong augmentation}
Strong augmentation ways can enrich the data set and improve the model performance easily. The methods in articles [29] [31] utilize color jittering, grayscale, gaussian blur, and cutout patches to augment data. And, cutout has a weak regularization effect.

\subsection{Weak augmentation}
Weak augmentation usually uses simple graphic transformation. Random horizontal flipping, random resizing and multi-scale are some regular weak augmentation ways[22] [23]. Mixmatch [19] expands the training data set by mixing up images from different classes randomly. Mix up has the problem of class ambiguity of mixture between backgrounds and objects.

\subsection{Hybrid augmentation}
To avoid above problems, both weak and strong augmentation are applied to a mini-batch unlabeled images in MUM method [25]. Besides, Instant-teaching [24] applies mosaic directly into pseudo label-based SSOD framework. STAC [26] explores different variants of transformation operations and determinates a group of effective combinations: 1) global color transformation; 2) global geometric transformation; 3) box-level transformation.

\section{Semi-supervised strategies}
As shown in Fig. 1, after data augmentation, the next step is to design a training framework to integrate information from both labeled and unlabeled images. Currently, SSOD methods follow four strategies. We will introduce some representative methods of these strategies, the flow charts of some classic methods are also presented.
\begin{figure*}[ht]%
\centering
\includegraphics[width=1.0\textwidth]{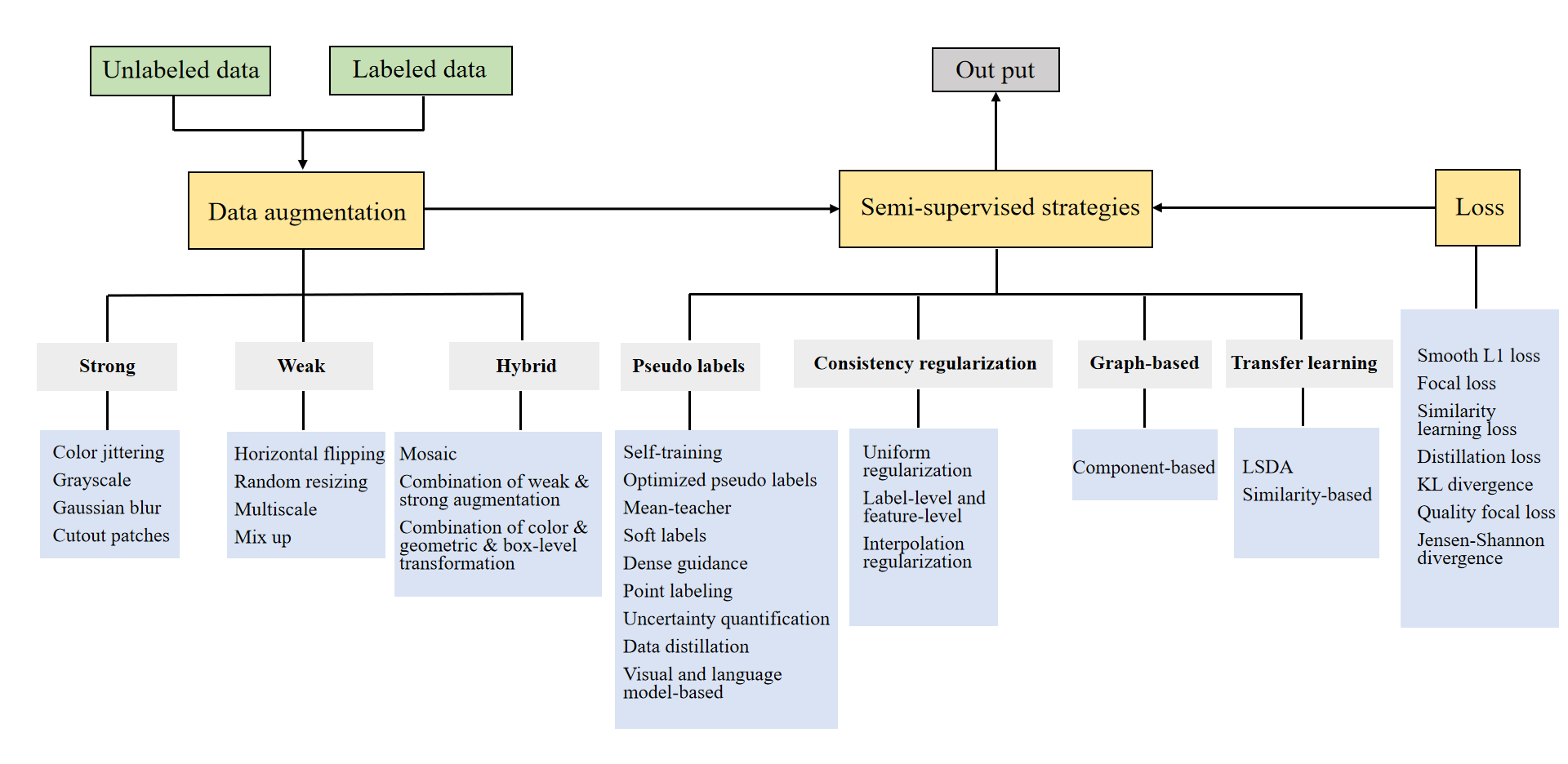}
\caption{Taxonomy of SSOD from the process perspective. }\label{fig:Fig1}
\end{figure*}
\subsection{Pseudo labels}
The first strategy is based on pseudo labels, as shown in Fig. 2, these methods estimate the pseudo labels for unlabeled images by using a pretrained model and then jointly train the model with both labeled and unlabeled data after augmentation. Most of them are based on two-stage anchor-based detector such as Faster-RCNN [52].
\begin{figure}[ht]%
\centering
\includegraphics[width=0.4\textwidth]{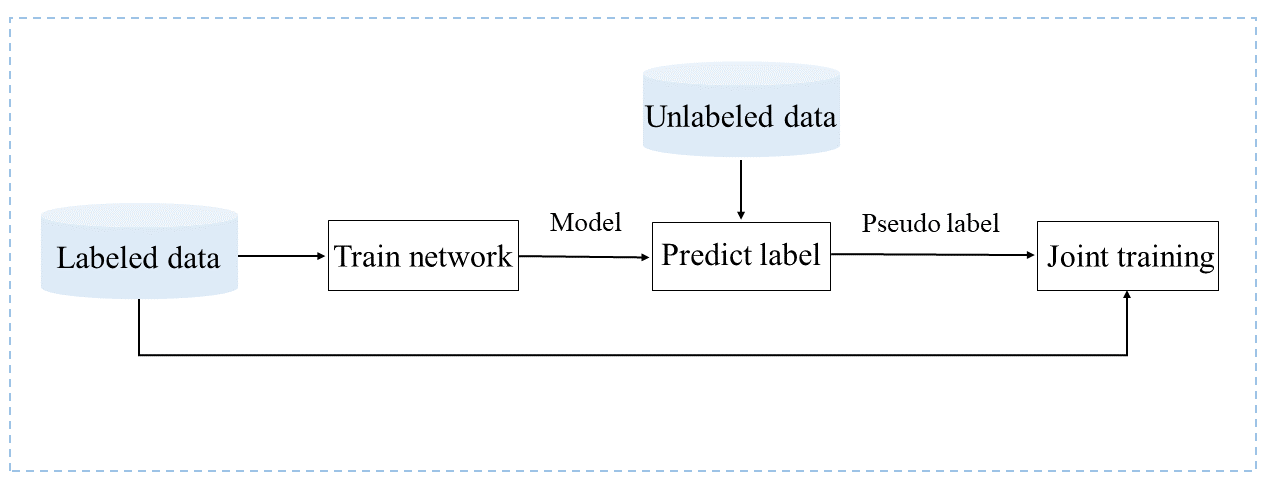}
\caption{The generation process of pseudo labels. }\label{fig:Fig2}
\end{figure}
\subsubsection{Self-training}
{Self-training uses labeled data to train a good teacher model which is used to predict the unlabeled data, and finally uses all data to train a student model. At present, many SSOD methods use the information of unlabeled samples by self-training for pseudo label prediction. It improves model performance by utilizing high-confidence samples with pseudo labels during the training process. 

The study [27] reveals that self-training improves upon pretraining and shows the generality and flexibility of self-training. STAC [21] proposes a SSOD algorithm based on hard pseudo label. The labeled data is used to train a teacher model which can predict unlabeled data, and it uses a threshold in order to select high confidence pseudo label and calculates the pseudo label with the unsupervised loss of the strong augmentation unlabeled data and the supervised loss of the labeled data. An overview of STAC method is shown in Fig. 3. 
\begin{figure}[ht]%
\centering
\includegraphics[width=0.4\textwidth]{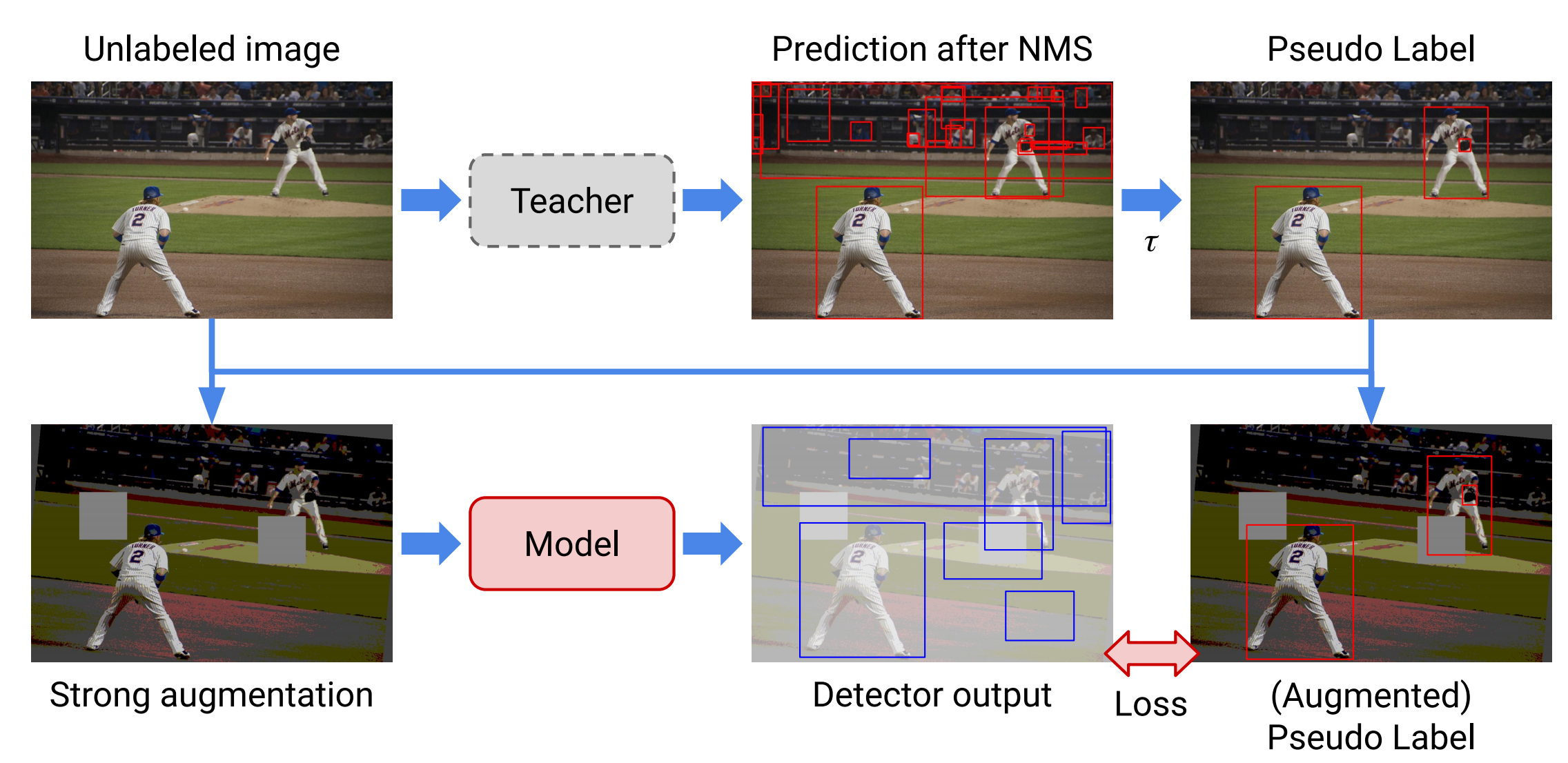}
\caption{Overview of STAC. }\label{fig:Fig3}
\end{figure}

In order to avoid ignoring the differences in the detection results in the same image during different training iterations and overfitting unlabeled data, as shown in Fig. 4, ISTM [28] proposes an interactive self-training model. On the one hand, it fuses the object detection results in different iterations with non-maximum suppression(NMS), on the other hand, it uses two region of interest(ROI) heads with different structures to estimate pseudo labels for each other.
\begin{figure}[ht]%
\centering
\includegraphics[width=0.4\textwidth]{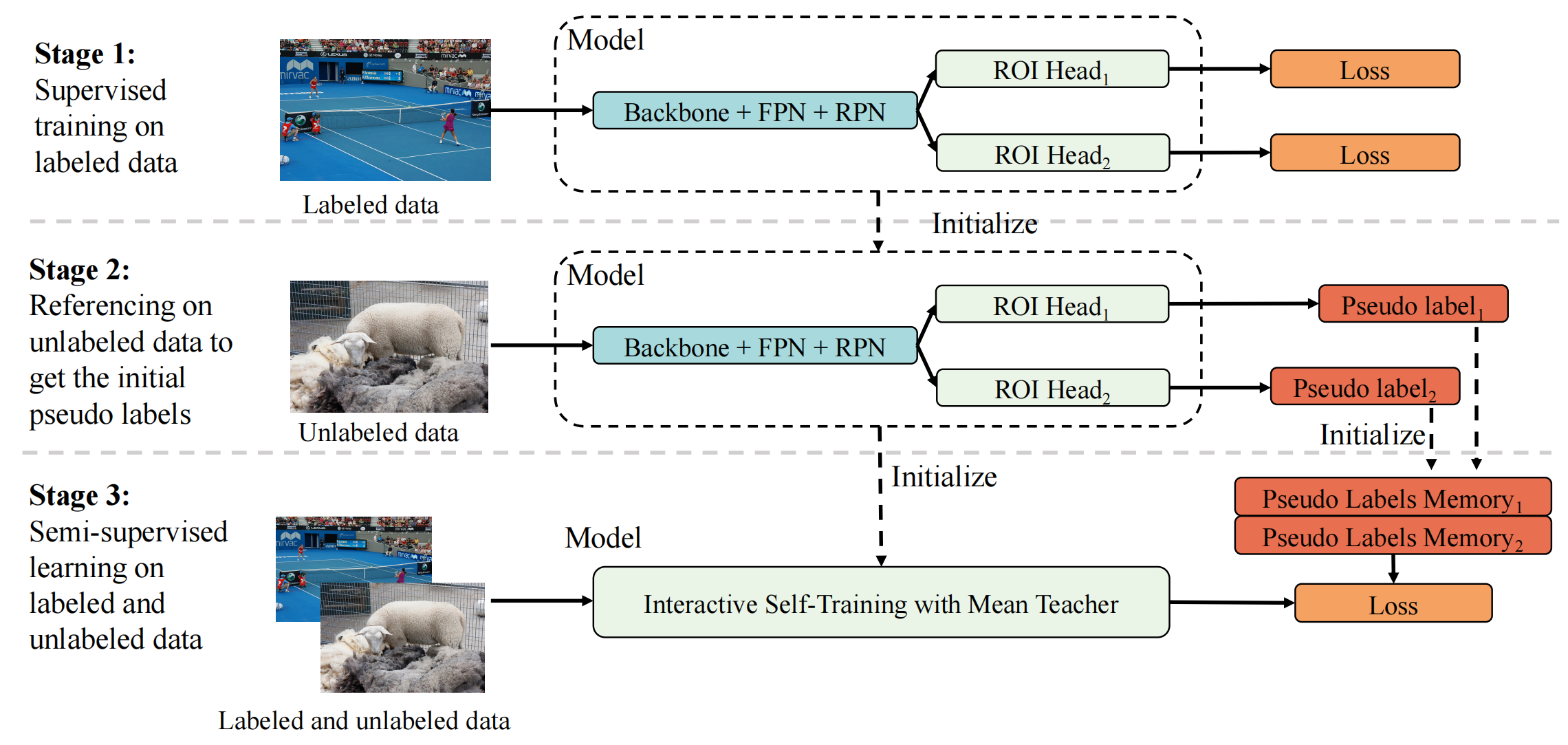}
\caption{Overview of ISTM. }\label{fig:Fig4}
\end{figure}

The class imbalance hinders the use of self-training. To address this problem, ACRST [23] proposes adaptive class-rebalancing self-training frame by rebalancing the training data with foreground instances extracted from the crop bank which stores abundant instance-level annotations.}

\subsubsection{Optimized pseudo labels}
{STAC generates only one-time pseudo labels, however, using the initial prediction of pseudo labels will limit the model accuracy improvement.

To alleviate the confirmation bias problem and improve the quality of pseudo labels, most approaches rectify the pseudo labels in training phase. As shown in Fig. 5, a two-stage framework is utilized by Unbiased Teacher [31], this method mitigates the overfitting problem by using pseudo labels to train region proposal network (RPN) and RoI head, solves the problem of pseudo label bias and improves the quality of pseudo label by using exponential moving average (EMA) and focal loss. 
\begin{figure}[ht]%
\centering
\includegraphics[width=0.4\textwidth]{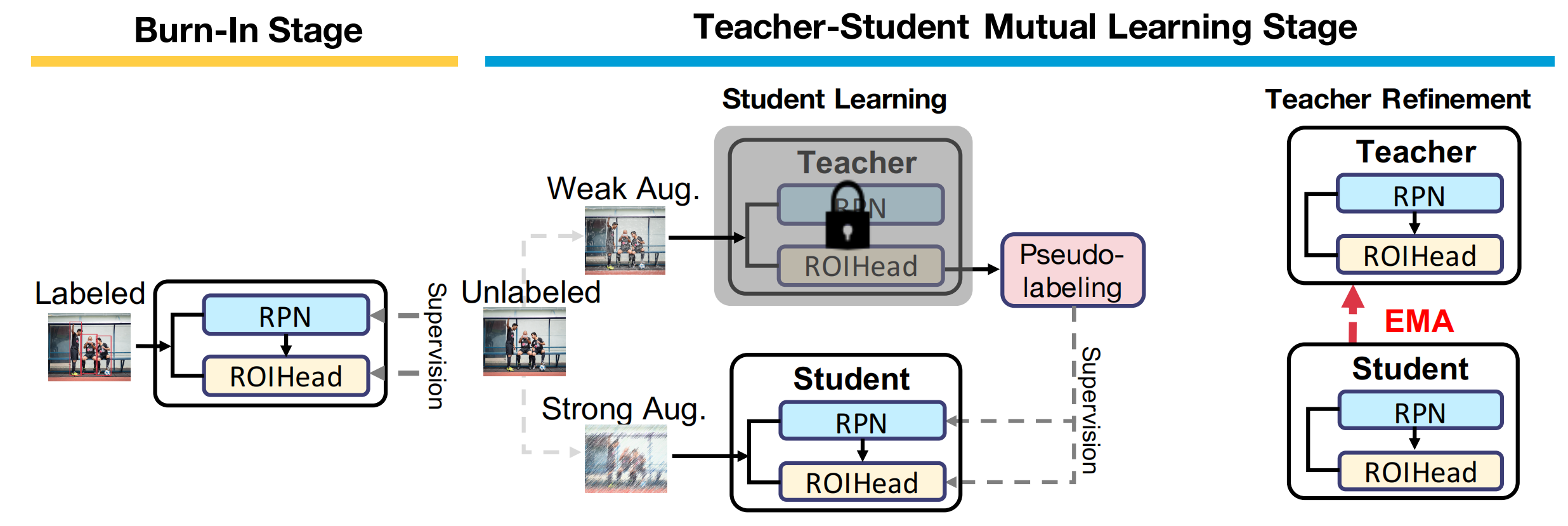}
\caption{Overview of Unbiased teacher. }\label{fig:Fig5}
\end{figure}
The confidence of the classification is used to screen the false tags of detection frame, which can't reflect the accuracy of localization. RPL [32] multiplies the mean value of classification confidence with the original detection classification confidence as an index which is used to reflect both classification accuracy and localization accuracy.

Based on the dual mining, to boost the tolerance to false pseudo labels, SIOD [33] mines latent instances from feature representation space by similarity-based pseudo label generating module (SPLG) and proposes pixel-level group contrastive Learning module (PGCL).

Cross Rectify [34] leverages the disagreements between detectors to discern the self-errors and refines the pseudo label quality by cross-rectifying mechanism. 

The MUM method [25] makes mixed input image tiles and reconstructs them in the feature space in SSOD framework.}

\subsubsection{Mean teacher}
{Mean teacher is widely used in these approaches [24] [30], it contains a teacher model whose weight is obtained from the EMA of the student model and a student model which needs to learn object generated from teacher. Soft teacher [30] proposes an end-to-end semi-supervised object detection method, in order to make full use of the information of the teacher model, the classification loss of the unlabeled bounding box is weighted by the classification score produced by the teacher network. Besides, it selects candidate boxes with box regression variance less than the threshold as pseudo labels for the better learning of box regression. The framework of Soft teacher is shown in Fig. 6.
\begin{figure}[ht]%
\centering
\includegraphics[width=0.4\textwidth]{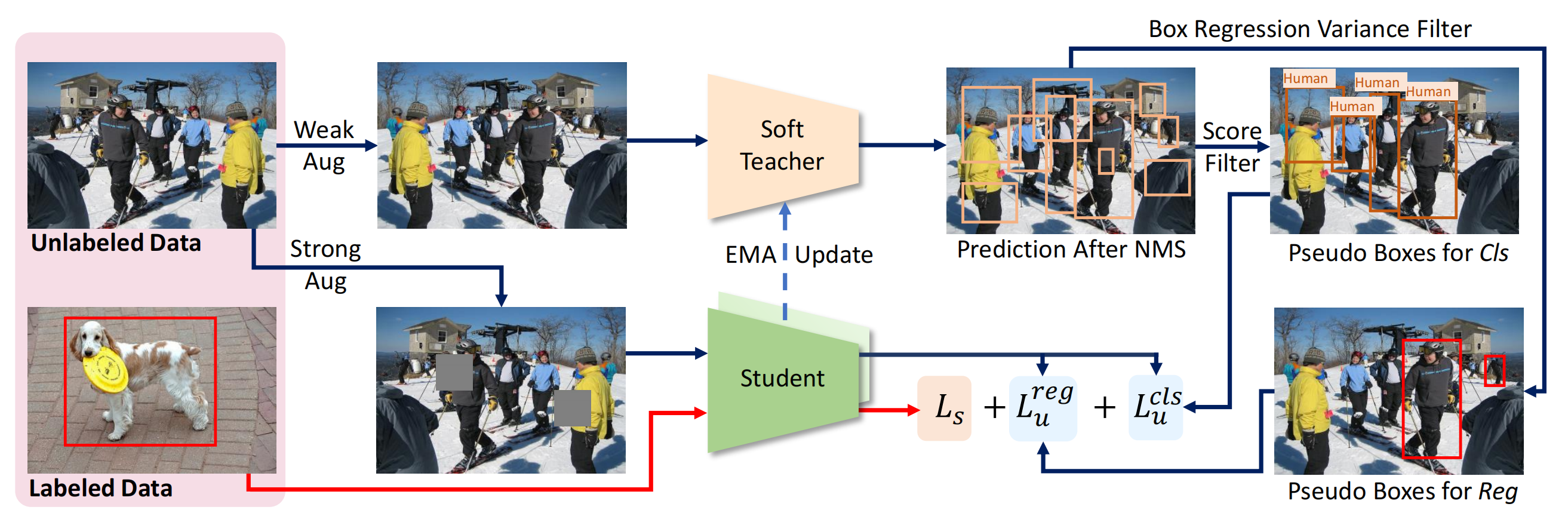}
\caption{Overview of Soft teacher.}\label{fig:Fig6}
\end{figure}

As shown in Fig. 7, Instant-teaching [24] proposes a co-rectify scheme which uses instant pseudo labeling with extended weak-strong data augmentations for teaching during each training iteration. 
\begin{figure}[ht]%
\centering
\includegraphics[width=0.4\textwidth]{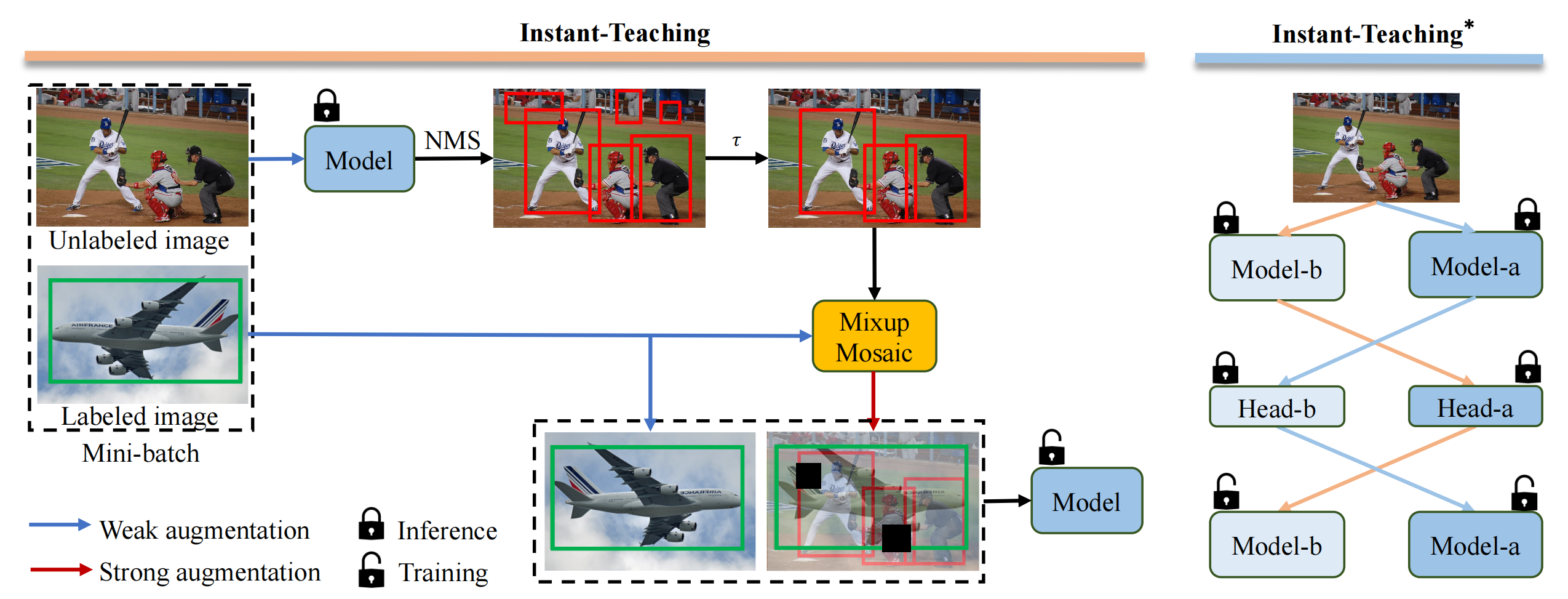}
\caption{Overview of Instant-teaching.}\label{fig:Fig7}
\end{figure}

Label matching [35] introduces a redistribution mean teacher, which leverages adaptive label-distribution-aware confidence thresholds to generate unbiased   pseudo labels to drive student learning.}

\subsubsection{Soft labels}
{Different from the STAC which adopts hard labels, soft labels are applied to Humble teacher [37] which gets the soft label targets from the predicted distribution of the class probabilities and offsets of all possible classes when the head is performing class-dependent bounding box regression. In order to provide more information, Humble teacher uses a large number of regional suggestions and soft pseudo labels as the training target of the student model. The framework of Humble teacher is shown in Fig. 8.}
\begin{figure}[ht]%
\centering
\includegraphics[width=0.4\textwidth]{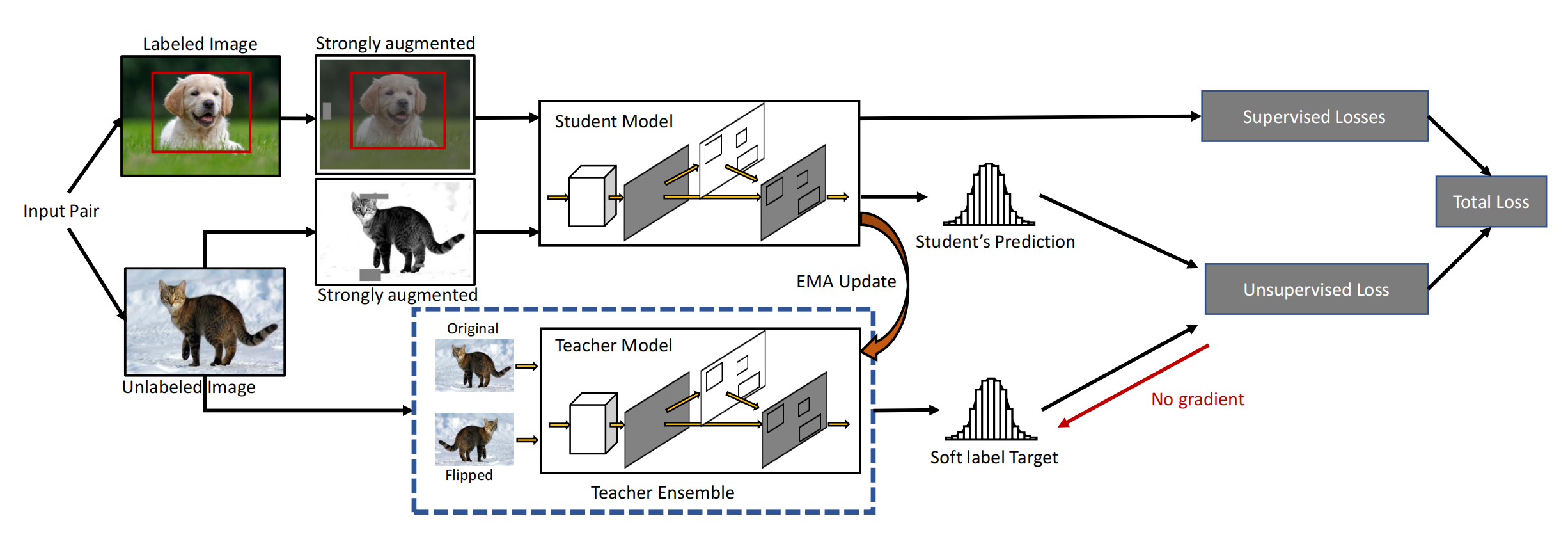}
\caption{Overview of Humble teacher.}\label{fig:Fig8}
\end{figure}

\subsubsection{Dense guidance-based}
{Dense Learning [22] puts forward adaptive filtering strategy and aggregated teacher for producing stable and precise pseudo-labels, besides, adopts uncertainty-consistency-regularization term among scales and shuffled patches for improving the generalization capability of the detector. The framework of Dense Learning is shown in Fig. 9.
\begin{figure}[ht]%
\centering
\includegraphics[width=0.4\textwidth]{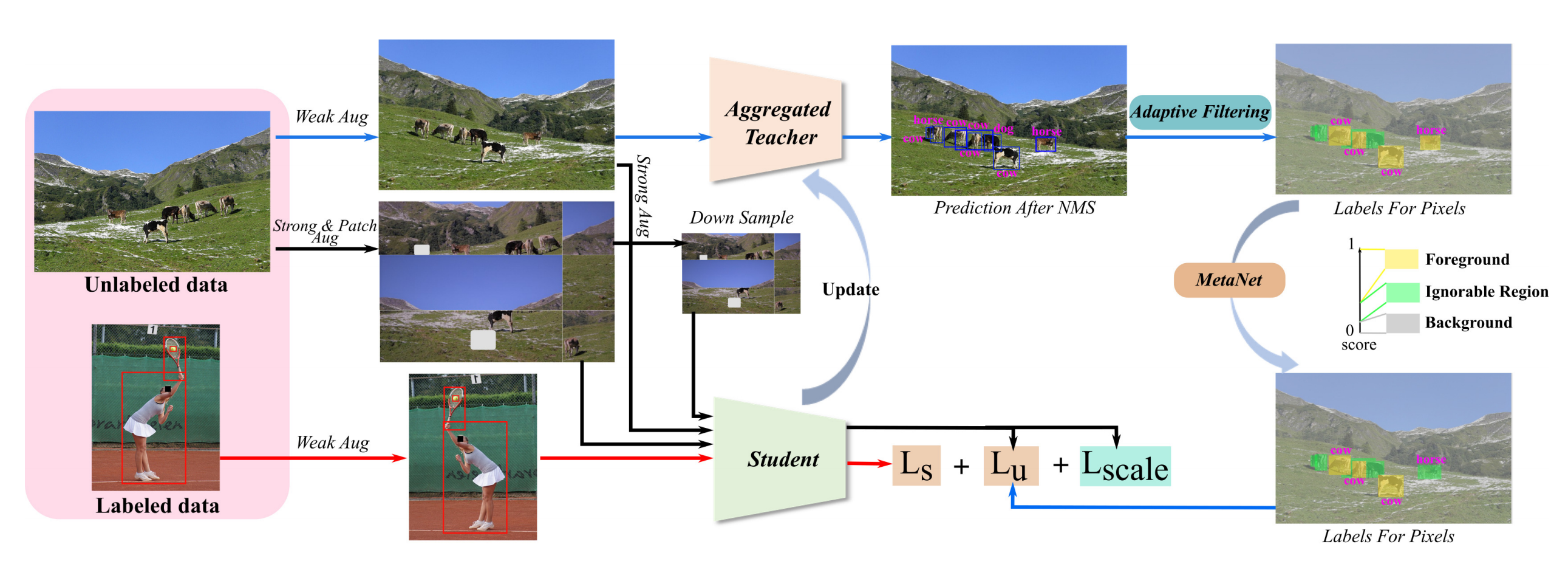}
\caption{Overview of Dense learning.}\label{fig:Fig9}
\end{figure}

Dense teacher [38] introduces a region selection technique to highlight the key information while suppressing the noise carried by dense labels. In order to replace sparse pseudo labels with more informative dense supervision, dense teacher guidance (DTG) [39] proposes a novel “dense-to-dense” paradigm which integrates DTG into the student training. It also introduces inverse NMS clustering and rank matching to make student be able to receive sufficient, informative, and dense guidance from the teacher, which naturally leads to better performance.}

\subsubsection{Point labeling}
{Point labeling can provide the location information of the instance, which greatly saves the labeling time. Under the unified architecture Omni-DETR [40], different types of weak labels can be leveraged to generate accurate pseudo labels, by a bipartite matching based filtering mechanism. Point DETR [41] extends DETR by adding a point encoder. As shown in Fig. 10, based on the classic R-CNN architecture, the point-to-box regressor Group R-CNN [42] proposes instance-level proposal grouping and instance-level representation learning by instance-aware feature enhancement and instance-aware parameter generation which aims to improve RPN recall and accomplish a one-to-one correspondence from the instance group to the instance box.}
\begin{figure}[ht]%
\centering
\includegraphics[width=0.4\textwidth]{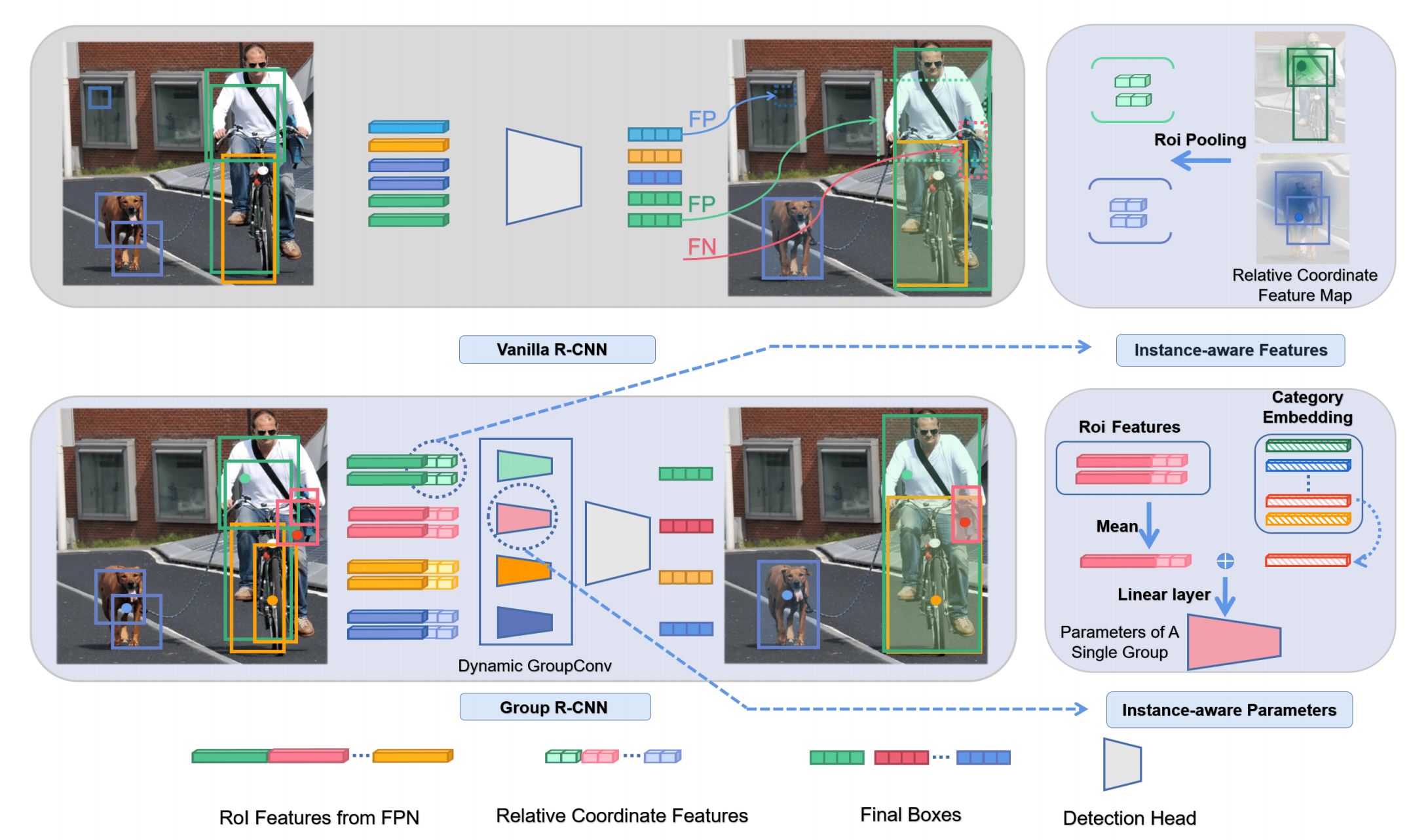}
\caption{Overview of Group R-CNN.}\label{fig:Fig10}
\end{figure}

\subsubsection{Uncertainty quantification}
{Label noise inherently exists in pseudo labels and brings about the uncertainty for SSOD training. There are several approaches to tackle this problem [43]-[45]. The method in literature [43] achieves noise-resistant learning by importing the region uncertainty quantification and promoting multipeak probability distribution output.

By utilizing the proposed uncertainty quantification as the soft target and facilitating multipeak probability distribution, as shown in Fig. 11, Combating noise [44] introduces uncertainty into semi-supervised learning.
\begin{figure}[ht]%
\centering
\includegraphics[width=0.4\textwidth]{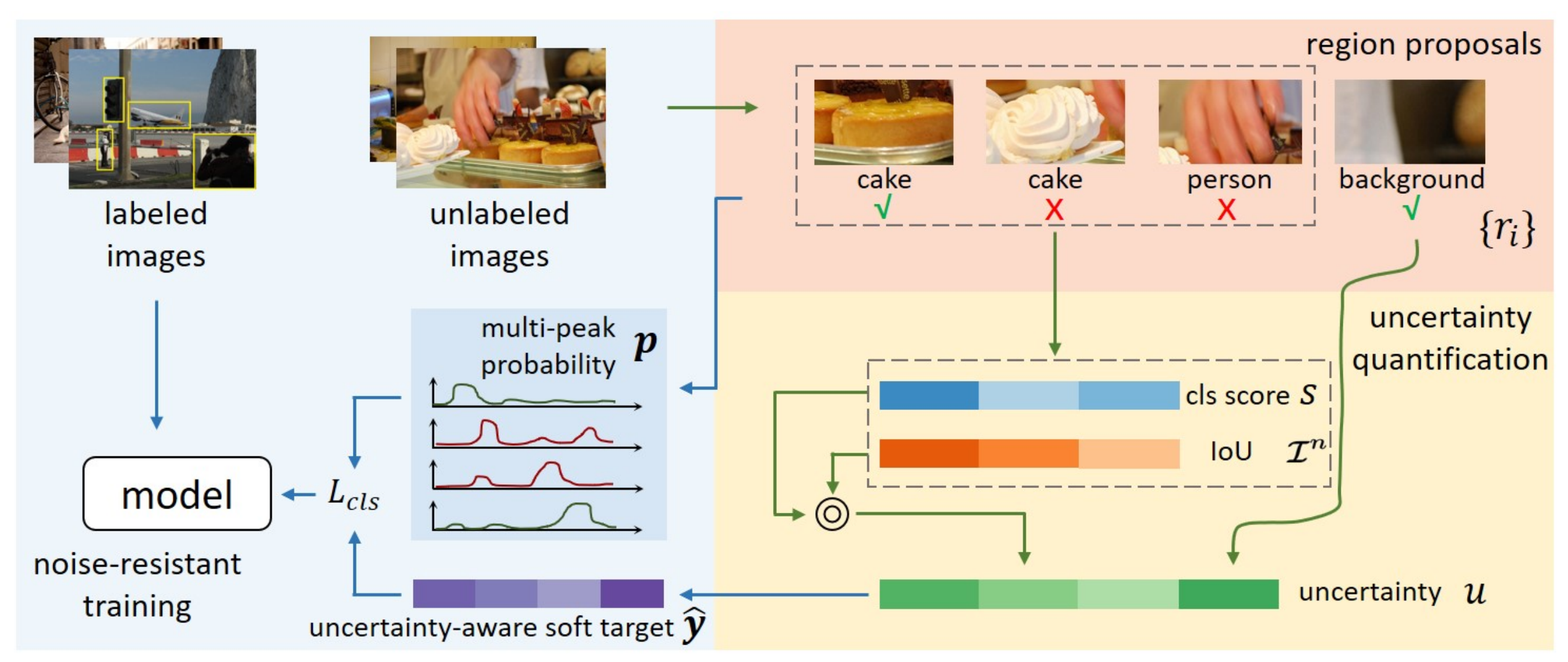}
\caption{Overall structure of Combating noise.}\label{fig:Fig11}
\end{figure}

In order to improve the filtering of the predicted bounding boxes and obtain higher quality on student training, an additional classification model for bounding box localization IL-Net [45] is introduced, utilizing a lightweight branch to predict the bounding box intersection over union (IoU) quality.

Due to the routine training-mining framework introduces various kinds of labelling noises easily, NOTE-RCNN [55] aims to handle noisy labels is proposed. As shown in Fig. 12, under the large number of image levels labels and a few seed box level annotations, the detector uses two classification heads and a distillation head to improve the mining precision, masks the negative sample loss and trains box regression heads only on seed annotations to eliminate the harm from inaccurate information.}
\begin{figure}[ht]%
\centering
\includegraphics[width=0.4\textwidth]{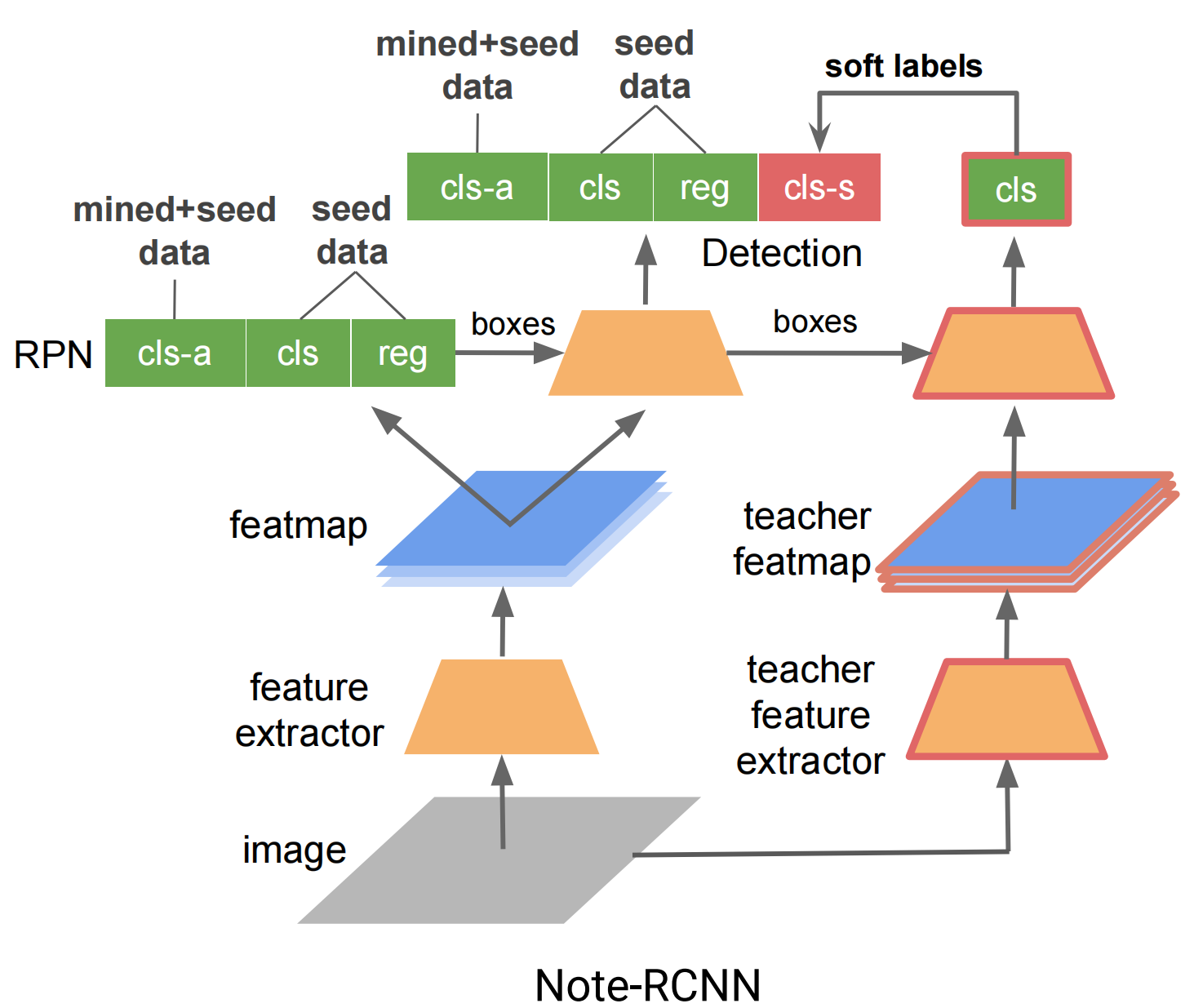}
\caption{The diagram of NOTE-RCNN.}\label{fig:Fig12}
\end{figure}

\subsubsection{Data distillation}
{The method in literature [36] proposes a self-distillation algorithm based on hint learning and ground truth bounded knowledge distillation to utilize purified data. The method in literature [29] is an omni-supervised setting which ensembles predictions from multiple transformations of unlabeled data by a single model, it also retrains the model on the union of the manually labeled data and automatically labeled data.}

\subsubsection{Visual and language model-based}
{Most prior works on SSOD only leveraged the small set of labeled data for generating pseudo labels, whereas the vision and language models are able to generate pseudo labels for both known and unknown categories.

As shown in Fig. 13, VL-PLM [46] starts with a generic training strategy for object detectors with the unlabeled data. In order to improve the pseudo label localization, it uses the class-agnostic proposal score and the repeated application of the RoI head, besides, provides better pseudo labels with judging the score of cropped regions with the visual and language model.
\begin{figure}[ht]%
\centering
\includegraphics[width=0.4\textwidth]{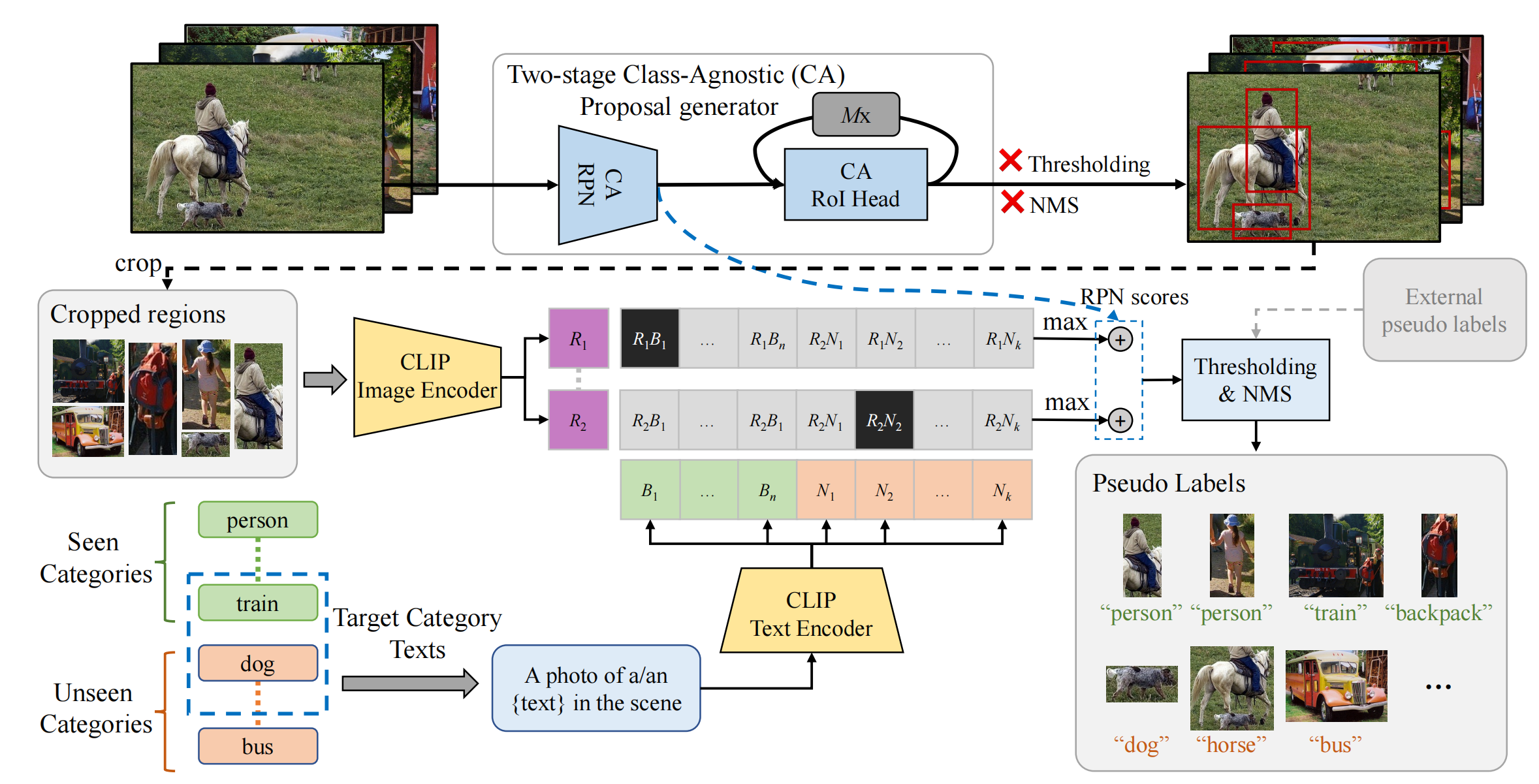}
\caption{Overall structure of VL-PLM.}\label{fig:Fig13}
\end{figure}

Prompt Det [47] is able to detect new categories without any manual annotation. As shown in Fig. 14, Prompt Det is divided into 3 stages, in the third stage, the base categories and novel categories are sent into the self-training network, the regional prompt learning is used to generate more accurate pseudo ground truth boxes.}
\begin{figure}[ht]%
\centering
\includegraphics[width=0.4\textwidth]{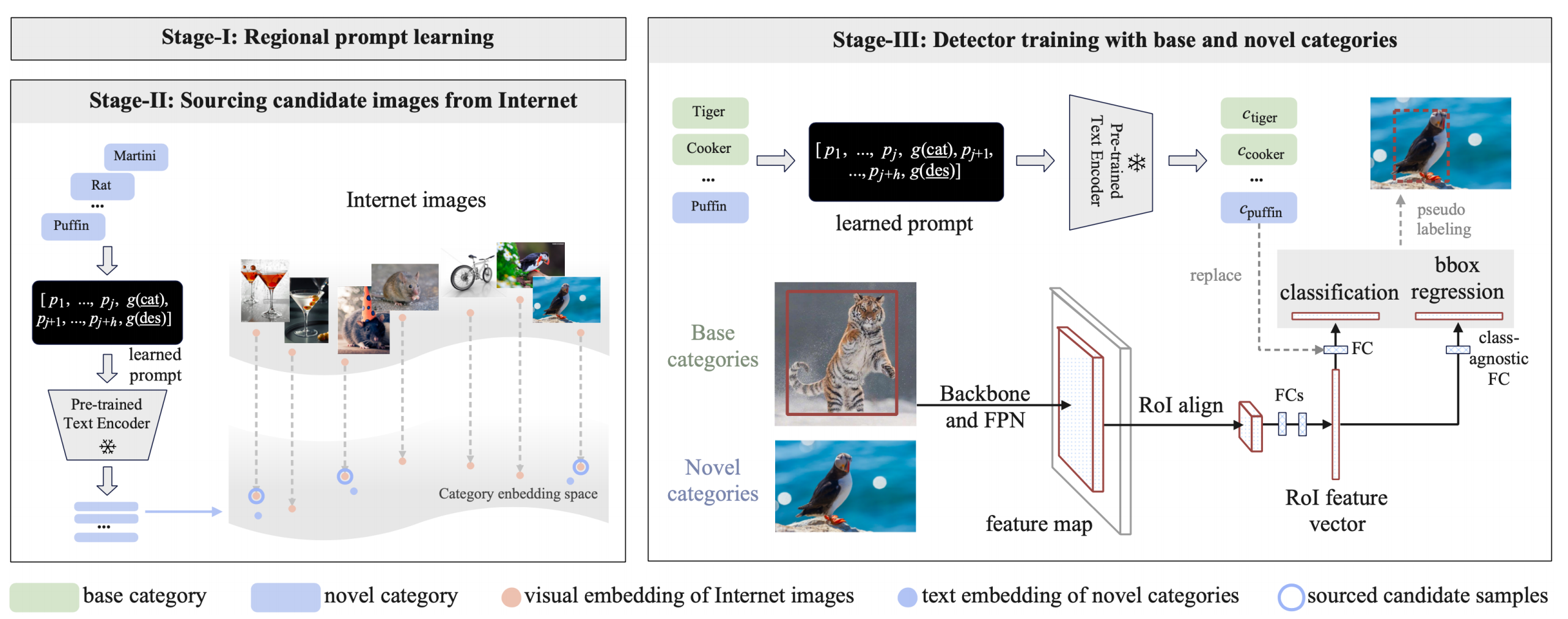}
\caption{Training process of Prompt det.}\label{fig:Fig14}
\end{figure}

\subsection{Consistent regularization}
The second strategy is based on consistency regularization, as shown in Fig. 15, these methods regularize the consistency of the outputs for the same unlabeled images under different forms of data augmentation. Most of them are based on two-stage anchor-based detector such as Faster-RCNN [52]. 
\begin{figure}[ht]%
\centering
\includegraphics[width=0.4\textwidth]{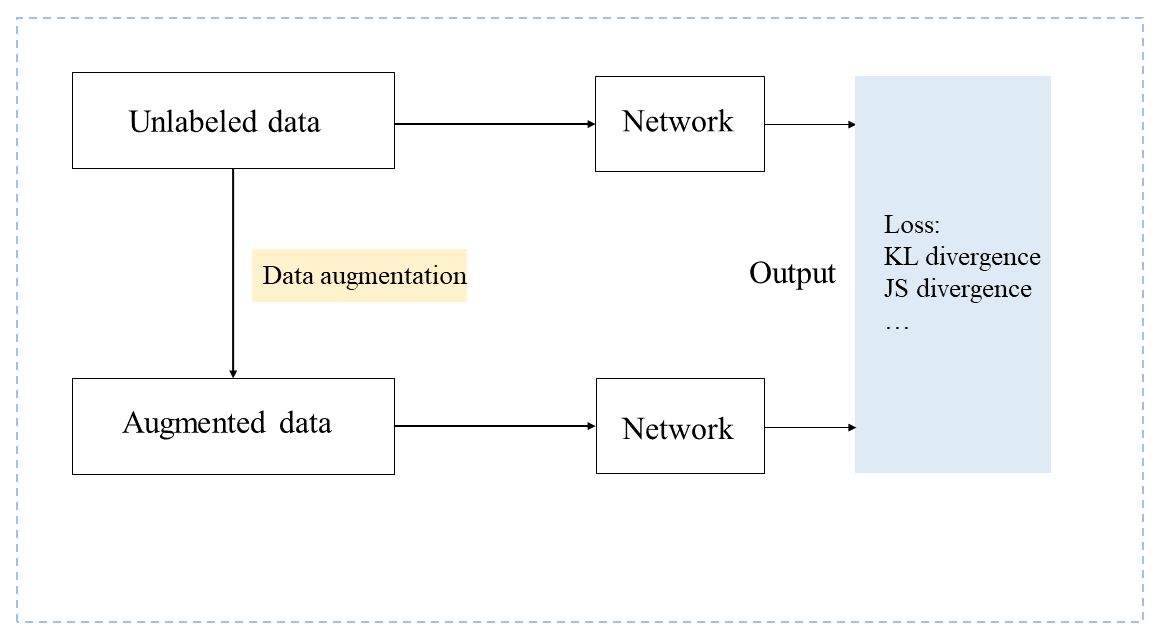}
\caption{The process of consistent regularization.}\label{fig:Fig15}
\end{figure}

CSD [48] is a typical semi-supervised object detection approach based on uniform regularization, which can work on both one-stage and two-stage detectors. In the first stage, the consistency loss of classification and localization is calculated according to the spatial position of the two images. In the second stage, the same set of RoI is generated by the same RPN for both images to extract features, and then the consistency loss is calculated. The overall structure of CSD is as shown in Fig. 16.
\begin{figure}[ht]%
\centering
\includegraphics[width=0.4\textwidth]{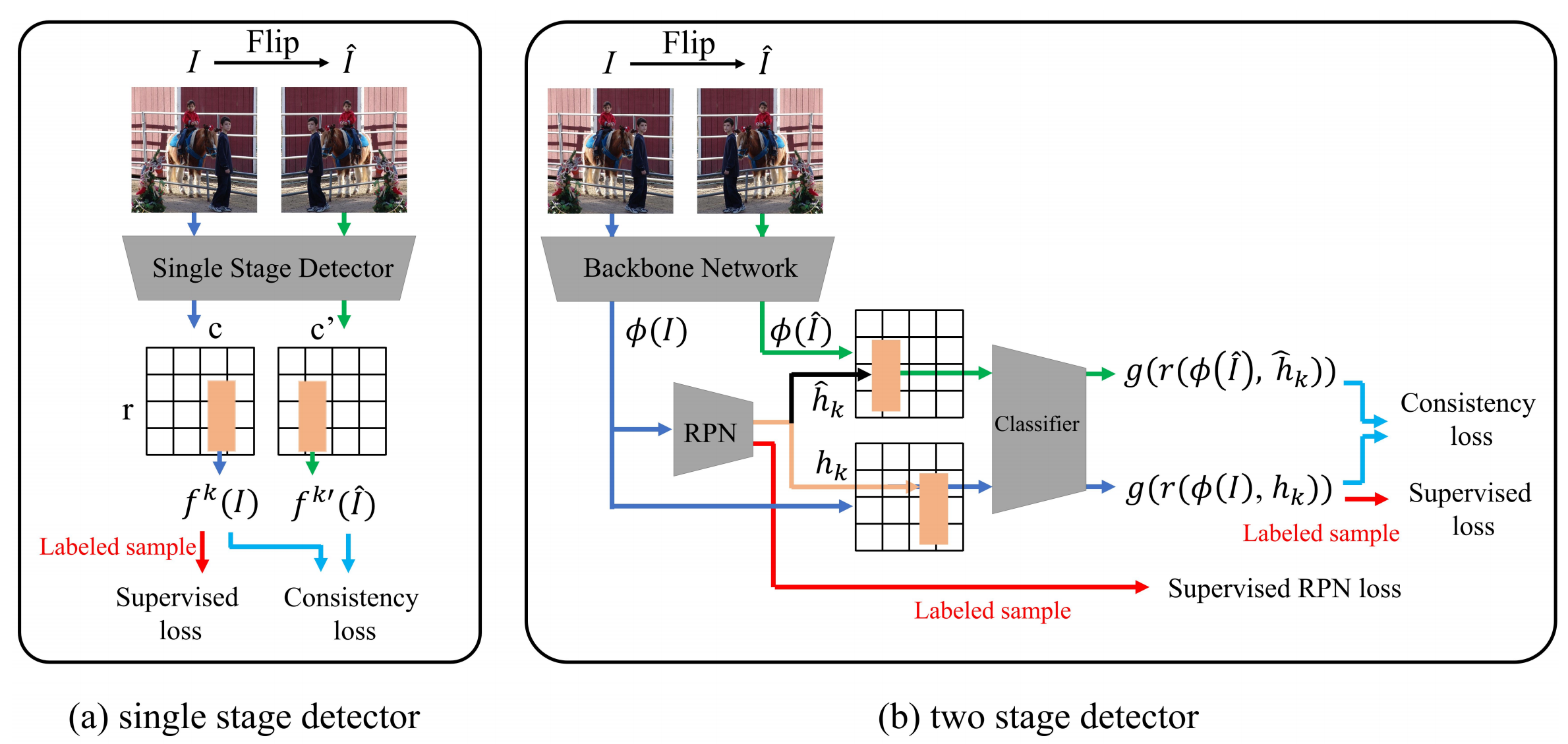}
\caption{Overall structure of CSD.}\label{fig:Fig16}
\end{figure}

In consistency training, PseCo [49] proposes a multiview scale invariant learning including label-level and feature-level consistency mechanisms, and feature consistency is achieved by aligning and shifting feature pyramids between two images with the same content but different scales.

ISD [50] applies interpolation regularization to object detection. It proposes different types of interpolation-based loss functions for each type in an unsupervised manner.

\subsection{Graph-based}
The third strategy is graph-based method. The labeled and unlabeled data points can be considered as nodes of a graph, and the objective is to propagate the labels from the labeled nodes to the unlabeled ones by utilizing the similarity of two nodes, which is reflected by how strong the edge between the two nodes. 

Graph-based method is an important semi-supervised learning branch in object tracking tasks, which can effectively utilize the comprehensive information of labeled and unlabeled samples. The graph-based method improves tracking accuracy by running graph-based semi-supervised classification methods on each graph independently, thus taking advantage of the intrinsic structural features of the sample set containing labeled as well as unlabeled samples.

The method in literature [53] proposes an online semi-supervised tracking way based on graph. Firstly, all samples are sampled based on the particle filter-based target motion model, and the candidate regions are obtained as unlabeled samples. Then, a component set is constructed for labeled and unlabeled samples, and a graph is constructed for each component set, on which the similarity between the unlabeled samples and labeled samples on the component is obtained by using the graph-based semi-supervised classification method independently. All similarity degrees are fused and unlabeled samples with the highest similarity are used as tracking results.

\subsection{Transfer learning-based}
It is always harder to get the object-level annotation (with category annotations and bounding box annotations) than image-level annotation (with category annotations). Therefore, how to transfer knowledge of existing categories with image annotations and object annotations to those categories without object level annotations is worth exploring. The fourth strategy is based on transfer learning, as shown in Fig. 17, it learns the difference between the two tasks and transfers knowledge of data without bounding box annotations from classifiers to detectors.
\begin{figure}[ht]%
\centering
\includegraphics[width=0.4\textwidth]{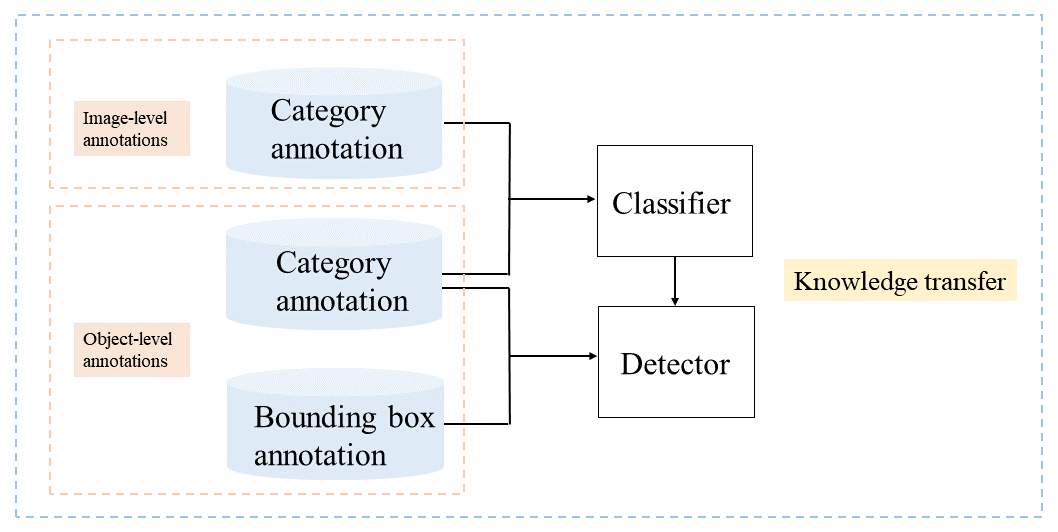}
\caption{The diagram of transfer learning.}\label{fig:Fig17}
\end{figure}

As shown in Fig. 18, a Large Scale Detection through Adaptation (LSDA) framework [54] is proposed,the algorithm uses the data with image-level annotation and data with object-level annotation to learn the classifier, then it converts the classification network into a detection network by using second kind of data,and finally gets adapted network by putting all the data into networks.
\begin{figure}[ht]%
\centering
\includegraphics[width=0.4\textwidth]{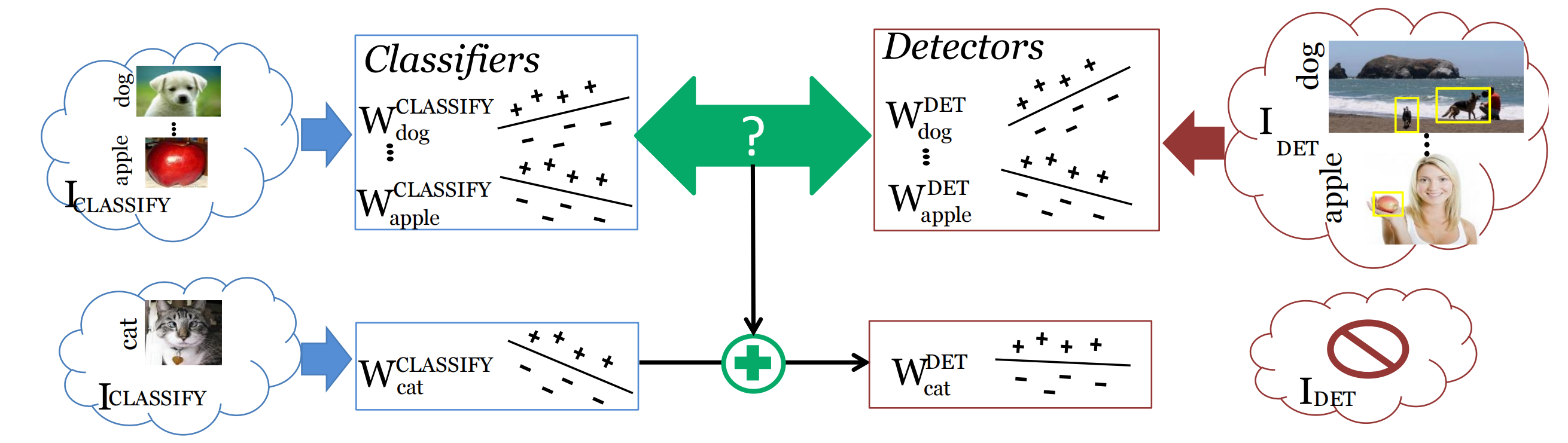}
\caption{The diagram of LSDA framework.}\label{fig:Fig18}
\end{figure}

On this basis, the method in literature [56] finds that visual similarity and semantic relatedness are complementary for the detection task. As shown in Fig. 19, similarity-based knowledge transfer model is proposed, it shows how knowledge about object similarities from both visual and semantic domains can be transferred to adapt an image classifier to an object detector in a semi-supervised setting. 
\begin{figure}[ht]%
\centering
\includegraphics[width=0.4\textwidth]{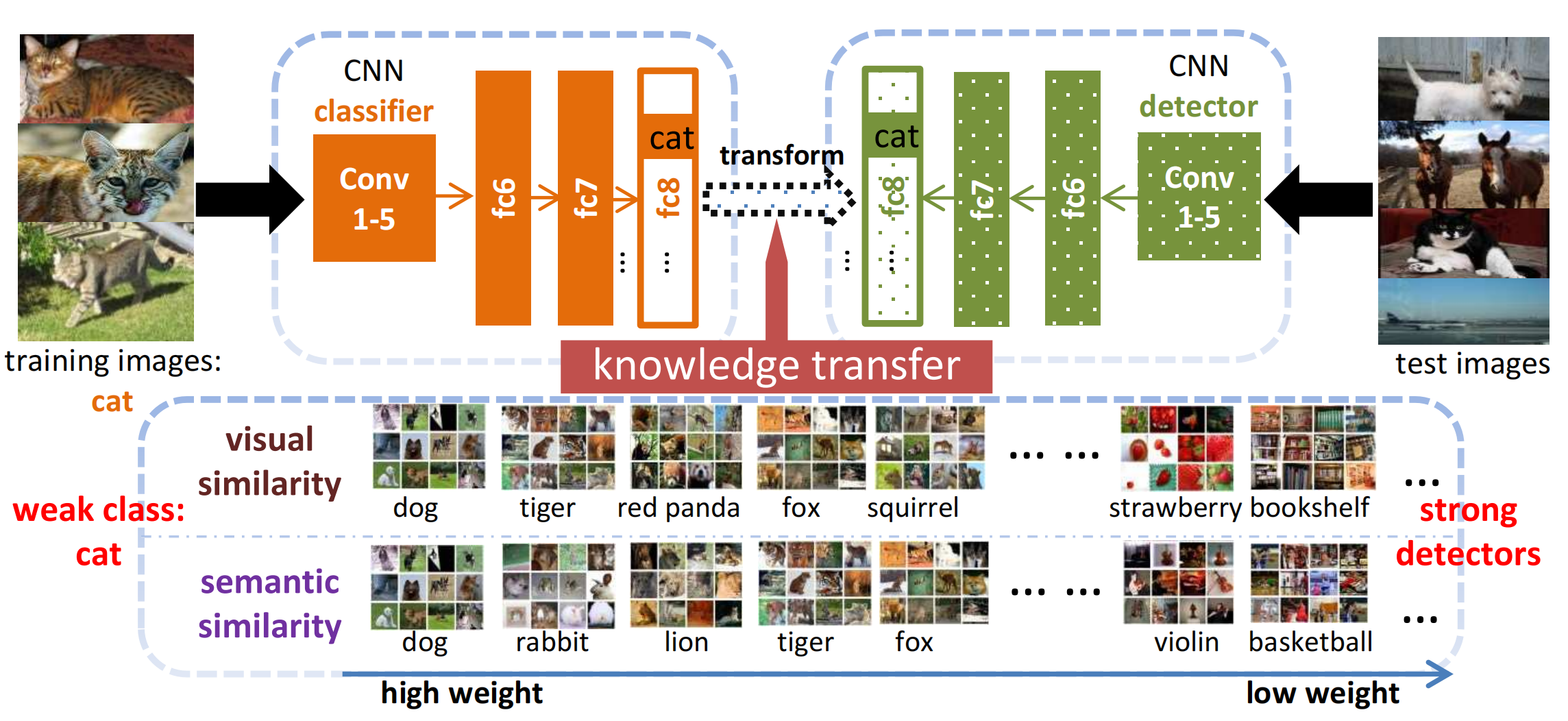}
\caption{The diagram of similarity-based knowledge transfer model.}\label{fig:Fig19}
\end{figure}

\section{Challenging setting}
In addition to the classical semi-supervised object detection methods mentioned above, some approaches have also been proposed in order to adapt challenging environments.

The VC learning [57] is assigned to each confusing sample such that they can safely contribute to the model optimization even without a concrete label. It also modifies the localization loss to allow high-quality boundaries for location regression. The pipeline of the VC learning is shown in Fig. 20.
\begin{figure}[ht]%
\centering
\includegraphics[width=0.4\textwidth]{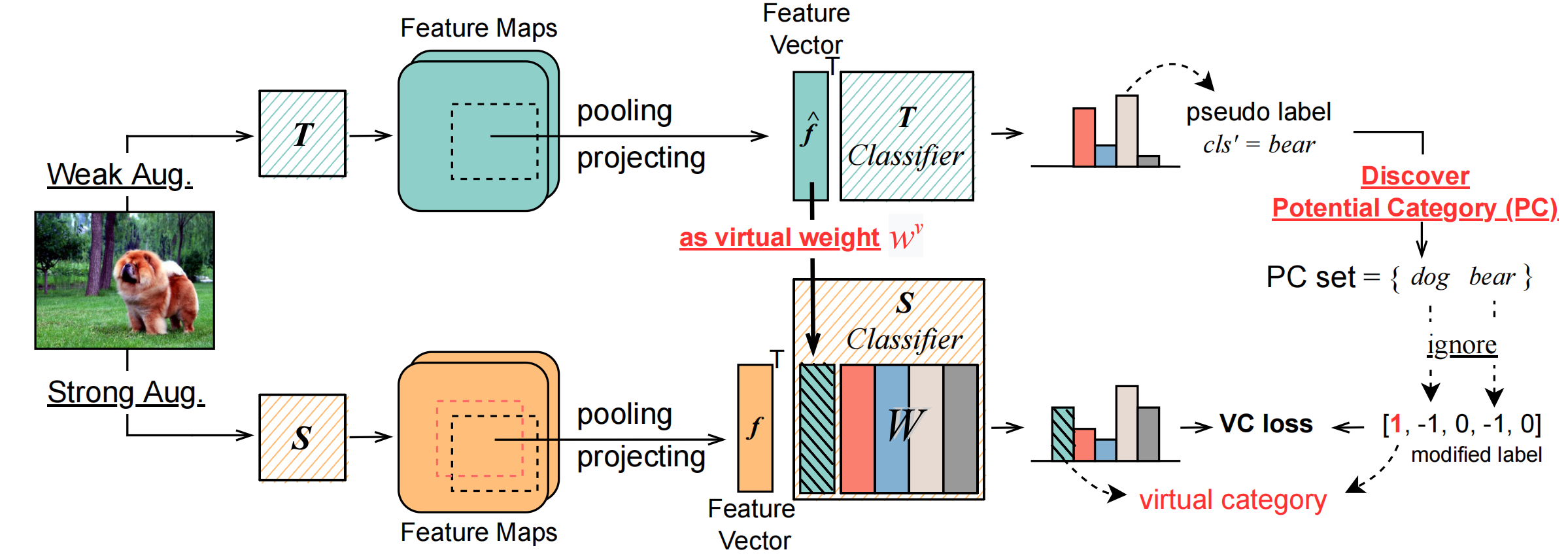}
\caption{The diagram of VC learning.}\label{fig:Fig20}
\end{figure}

The most SSOD approaches leverage unlabeled data to improve the performance of the detector. However, in fact, the unlabeled data always contain out-of-distribution (OOD) classes. In order to adapt open-set conditions, the open-Set semi-supervised object detection method [58] is proposed, the offline OOD detector based on a self-supervised vision transformer performs well on avoiding interference of pseudo labeling.

Semi-supervised long-tailed recognition is a new research field, the imbalanced data distribution and sample scarcity in its tail classes leads to low detection accuracy. The method in article [59] presents an alternate sampling framework which carries out supervised training and semi-supervised training in an alternate fashion together with model decoupling and different data sampling strategies.

These methods solve the problems in the task of semi-supervised object detection from different dimensions and have a strong reference for future work.

\section{Loss function}
According to Section I, after the data augmentation and semi-supervised strategies are used in network, the next step is to design suitable training losses for SSOD. The designed losses have a great impact on what SSOD can learn from the data. In this section, we introduce some widely-used loss functions.

In most SSOD methods, the overall loss is defined as the weighted sum of supervised loss and unsupervised loss, it can be formulated as follows:

\begin{equation}L=\mathrm{L}x+a\mathrm{L}u.\label{eq}\end{equation}

where Lx and Lu denote supervised loss of labeled images and unsupervised loss of unlabeled images respectively, a controls contribution of unsupervised loss. Both of them include classification loss and regression loss.

The losses for classification and localization are often instantiated as a weighted sum of a standard cross-entropy loss and a smooth L1 loss [35] [41] [43]. Other methods select loss functions with different characteristics.

\subsection{Smooth L1 loss}
The method in article [41] introduces the smooth L1 loss for both the supervised localization loss and consistency localization loss. The methods in articles [35] [43] also choose it for regression branch.

\subsection{Focal loss}
To alleviate imbalance issues, methods [17] [31] [39] choose focal loss on the unlabeled data. The method in article [45] sets focal loss for the new branch of IoU classification.

\subsection{Similarity learning loss}
In method [43], the similarity learning loss function is set up for supervisory signals. The loss of SIOD method [33] consists of a SPLG loss and a PGCL loss, the former focuses on determining latent instance feature similarity between labeled instances and unlabeled region, the latter is designed to boost the tolerance to false pseudo labels.

\subsection{Distillation loss}
Except the detection training loss, for better performance of multilayer hint for feature adaption, the method in article [37] also uses the knowledge distillation losses, Lkdc and Lkdr are utilized to quantify knowledge between student and teacher detectors in their classification and coordinates outputs.

\subsection{KL divergence}
Humble teacher [38] calculates KL divergence of the classification probability and bounding box regression output between the student RPN and teacher RPN, the same goes for the RoI phase. Combating Noise [44] utilizes the KL divergence for the classification branch.

\subsection{Quality focal loss}
DPL [38] adopts quality focal loss to conduct learning between dense pseudo labels and the student model’s predicting results because it needs to represent information in continuous values.

\subsection{Jensen-Shannon divergence}
In CSD method [48], Jensen-Shannon Divergence (JSD) is used as the consistency regularization loss for classification, the final loss is composed of the object detector’s classification loss, localization loss and consistency loss. The loss of ISD [50] is computed by Type-I loss and Type-II loss, the former uses jensen-shannon divergence as the consistency regularization loss, as for the latter, KL divergence and L2 loss are used as the classification and localization loss.
Besides, Dense learning [22] adopts the regularization loss for improving the generalization performance. The overall loss functions of DTG method [39] are classification loss, regression loss and rank matching loss which is adopted to optimize the distance between the score distribution of the teacher and student. omni-DETR [40] puts forward matching cost that encourages the selected predicted box to cover the ground truth point with small geometric distance and high confidence.

\section{Datasets and comparison results}
Using challenging datasets as benchmark is significant in object detection task, because they are able to make a standard comparison between different algorithms. There are now some quality detection datasets in the SSOD field. 

MS-COCO [3] dataset is a large-scale image dataset developed and maintained by Microsoft, and the corresponding tasks of dataset labeling types include object detection, key point detection, instance segmentation, stuff segmentation, panoramic segmentation of human key points, human density detection, etc. It contains more than 118k labeled images and 850k labeled object instances from 80 object categories for training. Besides, there are 123k unlabeled images that can be used for semi-supervised learning. The AP metrics of the SOTA methods on this dataset are summarized in the Table 1.

\begin{table}[ht]
\centering
\scriptsize
\caption{\textbf{Semi-supervised object detection results measured by AP50:95 on MS-COCO dataset. The performances are evaluated on the test set. 0.01, 0.02, 0.05 and 0.1 respectively represent the percentage of labeled data}}
\label{table1}
\begin{tabular}{|p{50pt}|p{30pt}|p{30pt}|p{30pt}|p{30pt}|}
\bottomrule
Method&
0.01&
0.02&
0.05&
0.1
\\
\hline
CSD&
10.51±0.06&
13.93±0.12&
18.63±0.07&
22.46±0.08
\\
STAC&
13.97±0.35&
18.25±0.25&
24.38±0.12&
28.64±0.21
\\
Unbiased Teacher&
20.75±0.12&
24.30±0.07&
28.27±0.11&
31.50±0.10
\\
Instant Teaching&
18.05±0.15&
22.45±0.30&
26.75±0.05&
30.40±0.05
\\
Soft Teacher&
20.46±0.39&
-&
30.74±0.08&
34.04±0.14
\\
Combating Noise&
18.41±0.10&
24.00±0.15&
28.96±0.29&
32.43±0.20
\\
Humble Teacher&
16.96±0.38&
21.72±0.24&
27.70±0.15&
31.61±0.28
\\
ISMT&
18.88±0.74&
22.43±0.56&
26.37±0.24&
30.52±0.52
\\
MUM&
21.88±0.12&
24.84±0.10&
28.52±0.09&
31.87±0.30
\\
Cross Rectify&
22.50±0.12&
27.60±0.07&
32.80±0.05&
36.30±0.07
\\
Label Match&
25.81±0.28&
-&
32.70±0.18&
35.49±0.17
\\
DTG-SSOD&
21.27±0.12&
26.84±0.25&
31.90±0.08&
35.92±0.26
\\
Dense Teacher&
22.38±0.31&
27.20±0.20&
33.01±0.14&
37.13±0.12
\\
Dense Learning&
22.03±0.28&
25.19±0.37&
30.87±0.24&
36.22±0.18
\\
\toprule
\end{tabular}
\end{table}

PASCAL-VOC [3] is made up of VOC07 and VOC12 which is often used for SSOD, there are 4 broad categories of datasets: vehicle, household, animal, and person. The trainval set of VOC07, containing 5011 images from 20 small object categories, it is used as a labeled training datasets. The trainval set of VOC12 contains 11,540 images, it is used as an unlabeled training datasets. The AP metrics of the SOTA methods on PASCAL-VOC dataset are summarized in the Table 2.

\begin{table}[ht]
\scriptsize
\centering
\caption{\textbf{Semi-supervised object detection results measured by AP50 and AP50:95 on PASCAL-VOC dataset. The performances are evaluated on the VOC07 test set}}
\label{table2}
\begin{tabular}{|p{50pt}|p{50pt}|p{50pt}|}
\bottomrule
Method&
Ap50&
Ap50:95
\\
\hline
CSD&
74.70&
- 
\\
STAC&
77.45&
44.64
\\
Unbiased Teacher&
77.37&
48.69
\\
Instant Teaching&
79.20&
50.00
\\
Combating Noise&
80.60&
-
\\
Humble Teacher&
80.94&
53.04
\\
ISMT&
77.23&
46.23
\\
MUM&
78.94&
50.22
\\
Omni-DETR&
-&
53.40
\\
Cross Rectify&
81.56&
-
\\
Label Match&
85.48&
55.11
\\
Dense Teacher&
79.89&
55.87
\\
Dense Learning&
80.70&
56.80
\\
\toprule
\end{tabular}
\label{tab1}
\end{table}

\section{Conclusion and Future directions}
Both the supervised algorithm and the semi-supervised algorithm can be applied to the object detection task. The supervised algorithm can achieve good performance, but also has certain limitations, it needs to learn a lot of labeled data and has high requirements on data quality. The semi-supervised algorithm introduced in this paper only requires a small amount of labeled data and a large amount of unlabeled data to improve the quality of the model, which saves the labeling cost in practice.
In this paper, we provide a complete review of newly proposed methodologies for semi-supervised object detection in the literature. We categorize the relevant methods based on their fundamental principles and describe their advantages and drawbacks, which may be of interest to practitioners as well as academic researchers. They are perhaps the most intuitive semi-supervised object detection methods, however, they still face many challenges. We make some interesting observations on possible future developments as follows.

\subsection{The accuracy of pseudo labels}
Self-training-based methods provide a flexibility for further development towards solving semi-supervised learning for object detection. The success of self-training models indicates the possibility of using a single self-training model to learn a representation of unlabeled data and building an intermediate label system to handle insufficient labeling problem in semi-supervision domain. However, the problems of confirmation bias and overly-used pseudo labels are ignored. Improved methods need to explore how to utilize unlabeled data more efficiently, whereas the confirmation bias problems are alleviated and the quality of pseudo labels are improved.

\subsection{Form of labels}
The pseudo label has proven to be effective in SSOD and has achieved state-of-the-art effect on benchmarks like MS-COCO and Pascal VOC. However, the process of generating pseudo labels requires several additional steps, such as NMS, thresholding, and label assignment. Dense guidance-based method is a pioneer work that makes the first step towards a more simple and effective form of pseudo labels. Soft labels coupled with a balanced number of teacher’s region proposals are the keys toward superior performance. Point labels-based method can also achieves a better cost-accuracy trade off. So it is useful to set a variety of label forms for obtaining more details of sample.

\subsection{Balance of categories}
The current SSOD approaches contribute to effectively improving the detection accuracy and the robustness against noisy samples under the balanced data. However, the need for large amounts of balanced labeled data in the training phase is hard to apply in real world scenarios. It is meaningful to take the balance of categories in unlabeled images into account and develop more new resource-friendly semi-supervised object detection methods.

\EOD

\end{document}